\documentclass{article}
\usepackage{ijcai25}

\usepackage{times}
\usepackage{soul}
\usepackage{url}
\usepackage[hidelinks]{hyperref}
\usepackage[utf8]{inputenc}
\usepackage[small]{caption}
\usepackage{graphicx}
\usepackage{amsmath}
\usepackage{amssymb}
\usepackage{amsthm}
\usepackage{booktabs}
\usepackage{algorithm}
\usepackage{algorithmic}
\usepackage[switch]{lineno}
\usepackage{subcaption}
\usepackage{xcolor}
\usepackage{longtable}
\usepackage{multirow}
\usepackage{booktabs}
\usepackage{array}
\usepackage{tikz}
\usepackage{hyperref}
\usetikzlibrary{positioning,shapes,shadows,arrows}
\usepackage{svg}
\usepackage{graphicx}
\usepackage{float}
\usepackage{booktabs} % For better table styling
\usepackage{amssymb}  % For \xmark
\usepackage{tikz}
\usetikzlibrary{arrows}
\usepackage{tabularx}

\linenumbers

\title{Large Language Models for Causal Discovery: Current Landscape and Future Directions}

\author{
Guangya Wan$^1$
\and
Yunsheng Lu$^2$ \and
Yuqi Wu$^3$\and
Mengxuan Hu$^1$\and
Sheng Li$^{1}$\\
\affiliations
$^1$School of Data Science, University of Virginia\\
$^2$Department of Statistics, University of Chicago\\
$^3$Department of Electrical and Computer Engineering, University of Alberta\\
\emails
\{wxr9et,qtq7su,shengli\}@virgnia.edu,
yunslu@uchicago.edu,
yuqi14@ualberta.ca,
}

\nolinenumbers  
\begin{document}

\maketitle
\begin{abstract}
Causal discovery (CD) and Large Language Models (LLMs) have emerged as transformative fields in artificial intelligence that have evolved largely independently. While CD specializes in uncovering cause-effect relationships from data, and LLMs excel at natural language processing and generation, their integration presents unique opportunities for advancing causal understanding. This survey examines how LLMs are transforming CD across three key dimensions: direct causal extraction from text, integration of domain knowledge into statistical methods, and refinement of causal structures. We systematically analyze approaches that leverage LLMs for CD tasks, highlighting their innovative use of metadata and natural language for causal inference. Our analysis reveals both LLMs' potential to enhance traditional CD methods and their current limitations as imperfect expert systems. We identify key research gaps, outline evaluation frameworks and benchmarks for LLM-based causal discovery, and advocate future research efforts for leveraging LLMs in causality research. As the first comprehensive examination of the synergy between LLMs and CD, this work lays the groundwork for future advances in the field.

\end{abstract}

\begin{figure*}[!t]
\centering
\begin{tikzpicture}[
every node/.style={
draw=red,
rounded corners,
align=left,
minimum height=0.5cm,
inner sep=5pt,
inner xsep=5pt,
font=\small
},
level/.style={
level distance=4cm,
edge from parent/.style={
draw,
-,
thin
}
}
]
% Root node
\node (root) [text width=4cm,rotate=270] at (-17,0) {LLMs for Causal Discovery};

% Level 1 nodes - increased vertical spread
\node (background) [text width=3cm] at (-13.5,5) {Background (\S \ref{background})};
\node (methods) [text width=3cm] at (-13.5,2.5) {Methodology (\S\ref{llm4causal})};
\node (eval) [text width=3cm] at (-13.5,-1.5) {Evaluations (\S\ref{Eval})};
\node (future) [text width=3cm] at (-13.5,-4.5) {Future Directions (\S\ref{Future})};

% Level 2 nodes - Methods children - increased spacing
\node (direct) [text width=3.5cm] at (-9,4.5) {As Direct Inference (\S\ref{DI})};
\node (posthoc) [text width=3.5cm] at (-9,2.5) {As Post Refinement (\S\ref{Posterior})};
\node (prior) [text width=3.7cm] at (-8.8,0.5) {As Prior Knowledge (\S\ref{Prior})};

% Level 2 nodes - Evaluation children - adjusted positions
\node (bench) [text width=4.5cm] at (-8.8,-1.5) {Benchmarks and Datasets (\S\ref{Data})};
\node (apps) [text width=2.6cm] at (-9,-4) {Applications (\S\ref{Apps})};

% Level 3 nodes - Citation boxes with adjusted vertical positions
\node (direct-cite) [text width=6cm, fill=blue!10] at (-3,4.5) {\cite{willig2022foundation}, \cite{ban2023causal}, \cite{kıcıman2023causal}, \cite{vashishtha2023causalinferenceusingllmguided}, \cite{jiralerspong2024efficient}, \cite{le2024multiagentcausaldiscoveryusing}, \cite{long2024large}, \cite{sokolov24}, \cite{zhang2024causalgraphdiscoveryretrievalaugmented}};

\node (posthoc-cite) [text width=6cm, fill=yellow!10] at (-3,2.5) {\cite{ban2023causal}, \cite{long2023causal}, \cite{abdulaal2024causal}, \cite{takayama2024integrating}};

\node (prior-cite) [text width=6cm, fill=green!10] at (-3,0.5) {\cite{ban2023query}, \cite{chen2023mitigatingpriorerrorscausal}, \cite{cohrs2023large}, \cite{darvariu2024large}, \cite{kampani2024llminitializeddifferentiablecausaldiscovery}};

\node (bench-cite) [text width=6cm, fill=orange!10] at (-3,-1.5) {\cite{heindorf2020causenet}, \cite{huang2021benchmarking}, \cite{jin2023cladder}, \cite{jin2023large}, \cite{kıcıman2023causal}, \cite{tu2023causaldiscovery}, \cite{abdulaal2024causal}, \cite{liu2024discoveryhiddenworldlarge}, \cite{zhou2024causalbenchcomprehensivebenchmarkcausal}};

\node (apps-cite) [text width=6cm, fill=purple!10] at (-3,-4) {\cite{10.1145/3698587.3701384}, \cite{tu2023causaldiscovery}, \cite{abdulaal2024causal}, \cite{afonja2024llm4grndiscoveringcausalgene}, \cite{gkountouras2024languageagentsmeetcausality}, \cite{jiang2024llmcausal}, \cite{liu2024discoveryhiddenworldlarge}, \cite{shen2024exploringmultimodalintegrationtoolaugmented}, \cite{sokolov24}, \cite{cohrs2025large}};

% Main horizontal line from root
\draw [->] (root) -- (-15.5,0);
\draw [-] (-15.5,-4.5) -- (-15.5,5);

% Connect to level 1 nodes
\draw [-] (-15.5,5) -- (background);
\draw [-] (-15.5,2.5) -- (methods);
\draw [-] (-15.5,-1.5) -- (eval);
\draw [-] (-15.5,-4.5) -- (future);

% Vertical line for methods
\draw [-] (-11.5,4.5) -- (-11.5,0.5);
% Connect methods to its children
\draw [-] (methods) -- (-11.5,2.5);
\draw [-] (-11.5,4.5) -- (direct);
\draw [-] (-11.5,2.5) -- (posthoc);
\draw [-] (-11.5,0.5) -- (prior);

% Vertical line for evaluation - adjusted
\draw [-] (-11.5,-1.5) -- (-11.5,-4);
% Connect evaluation children
\draw [-] (-11.5,-1.5) -- (bench);
\draw [-] (-11.5,-4) -- (apps);
% Connect eval to its vertical line
\draw [-] (eval) -- (-11.5,-1.5);

% Connect Methods subsections to their citation boxes with straight lines
\draw [-] (direct) -- (direct-cite);
\draw [-] (posthoc) -- (posthoc-cite);
\draw [-] (prior) -- (prior-cite);
% Connect Evaluation subsections to their citation boxes with straight lines
\draw [-] (bench) -- (bench-cite);
\draw [-] (apps) -- (apps-cite);

\end{tikzpicture}
\caption{Overview of LLM-based causal discovery methods, categorized by their role: direct evaluation, prior knowledge augmentation, and post-hoc refinement. Evaluation strategies and future directions are also outlined.}
\label{fig:taxonomy}
\end{figure*}

\section{Introduction}

Uncovering causal relationships—understanding why things happen—is fundamental to scientific discovery and informed decision-making across diverse domains. From discovering the causes of diseases and developing effective treatments to optimizing complex systems like city traffic flow or global supply chains, knowing why something occurs is crucial for effective intervention \cite{kuang2020causal}. Causal Discovery (CD) has long relied on two pillars: statistical methods for data analysis and domain experts for knowledge integration \cite{pearl2009causal}. While domain experts provide invaluable insights drawn from years of experience and deep understanding, their involvement often creates bottlenecks in the discovery process. Consulting experts is time-consuming, expensive, and inherently limited by human availability and potential biases. Meanwhile, Statistical CD methods (SCD) \cite{shimizu2006linear,shimizu2011directlingam,ScutariDenis2014,zheng2018dagstearscontinuousoptimization}, while mathematically rigorous, often fall short in real-world scenarios. They typically demand vast amounts of high-quality data, which is often unavailable or expensive to acquire. Furthermore, they struggle to disentangle complex temporal dynamics inherent in many real-world systems, where causes and effects unfold over time and influence each other in intricate ways \cite{ban2023causal}. Specifically, these methods frequently produce multiple, equally plausible causal explanations, traditionally requiring expert intervention to resolve these ambiguities.

Large Language Models (LLMs) offer a transformative solution to the challenges of causal discovery, potentially acting as scalable and generalized "meta-experts" \cite{kıcıman2023causal}.  Their ability to process and synthesize massive amounts of text—effectively distilling knowledge from countless documents, research papers, and expert opinions—makes them powerful tools for automating expert-level reasoning \cite{zhao2023survey}.  Unlike traditional domain experts with specialized knowledge, LLMs can provide broad expertise across multiple domains concurrently. They can integrate information from diverse sources to identify potential causal relationships that might be overlooked by statistical methods or limited human consultation.  For example, when analyzing urban traffic, an LLM could rapidly synthesize insights from thousands of traffic engineering papers, weather studies, and city planning documents to suggest previously unrecognized causal factors—a task that could take human experts weeks or months.

The integration of LLMs into CD represents a paradigm shift from both purely statistical approaches and traditional expert-dependent methods, manifesting in three primary ways: (1) LLMs can directly infer causal graphs or subgraph structures from natural language descriptions and domain knowledge \cite{jin2023large}, effectively automating the initial expert hypothesis generation phase. (2) LLMs can function as posterior correction mechanisms, validating and refining causal relationships identified by SCD methods against their extensive knowledge base \cite{long2024large}, similar to how experts would review and adjust statistical findings. (3) They can serve as comprehensive prior information sources for traditional SCD algorithms, providing domain knowledge and contextual constraints before statistical analysis \cite{takayama2024integrating}. This systematic replacement of human expert involvement with LLM-based automation not only accelerates the discovery process but also makes sophisticated causal analysis accessible to a broader range of researchers and practitioners without domain expertise.

Several surveys have examined how large language models (LLMs) relate to causality. However, there is still a lack of detailed survey specifically on \textit{causal discovery} in the era of LLMs.
Traditional causal discovery surveys like \cite{glymour2019review} extensively explore connections with machine learning and deep learning approaches but lacks discussion with the emergence of LLMs in this domain. \cite{zhao2023survey} pioneered the discussion of LLMs in causal reasoning but primarily focuses on broader tasks such as counterfactual reasoning, cause attribution, and causal effect estimation. More recent work by \cite{zhang2024causalgraphdiscoveryretrievalaugmented} examines causality in LLMs but provides a limited analysis of how these models fundamentally transform causal discovery methods. Similarly, \cite{yu2024improvingcausalreasoninglarge} offers a comprehensive review of improving LLMs' causal reasoning capabilities but lacks specific focus on causal discovery tasks and their integration with traditional causal theory. To address this gap in literature, our survey examines the emerging intersection of LLMs and CD, providing a systematic framework for understanding their integration. We begin in Section 2 with background on both LLMs and CD, bridging knowledge gaps for researchers from either field. Section 3 analyzes how LLMs can enhance CD through direct inference, prior knowledge integration, and structural refinement. Section 4 evaluates benchmark datasets and showcases applications across diverse domains, from healthcare to social sciences. Section 5 concludes by examining current limitations and identifying promising research directions that could advance this rapidly evolving field.

\section{Background}
\label{background}
This section lays the groundwork for understanding how Large Language Models (LLMs) can revolutionize causal analysis.  We review fundamental concepts and methodologies of causal inference and discovery, including essential notations and definitions in causal discovery to prepare readers better understand and subsequent technical sections.

\subsection{Graphical Models}
\textbf{Directed Acyclic Graph (DAG)}: A directed acyclic graph is an ordered pair \( \mathcal{G} = (\mathcal{V}, \mathcal{E}) \), where \( \mathcal{V} \) is a finite set of vertices and \( \mathcal{E} \subseteq \mathcal{V} \times \mathcal{V} \) is a set of directed edges such that for any \( (X, Y) \in \mathcal{E} \), the relation \( X \to Y \) holds. Additionally, \( \mathcal{G} \) contains no directed cycles, i.e., there does not exist a sequence of distinct nodes \( X_1, X_2, \dots, X_k \) such that \( X_1 \to X_2 \to \dots \to X_k \to X_1 \).

% \noindent \textbf{Bayesian Network (BN)}: A DAG $(\mathcal{V}, \mathcal{E})$ is a Bayesian network for a probability measure $\mathbb{P}(\textbf{X})$ defined over a set of random variables $\mathbf{\textbf{X}}$, if \( \mathcal{V} \) represents the set of random variables $\mathbf{X}$, and $\mathbb{P}(X)$ factorizes according to \( \mathcal{G} \), that is:
%     \begin{equation}\label{recursivefac}
%        \mathbb{P}_\mathbf{X} = \prod_{i=1}^{n} \mathbb{P}(X_i \mid \text{Pa}(X_i)), \quad  \text{Pa}(X_i) = \text{Parents of } X_i
%     \end{equation}
    
\begin{figure}[t]
    \centering
    % First DAG: X1 -> X2 -> X3
    \begin{tikzpicture}[->, >=stealth, node distance=1cm]
        \node (X1) at (0,0) {$X_1$};
        \node (X2) at (1,0) {$X_2$};
        \node (X3) at (2,0) {$X_3$};

        \draw (X1) -- (X2);
        \draw (X2) -- (X3);
    \end{tikzpicture}
    \hspace{1cm}
    % Second DAG: X1 <- X2 <- X3
    \begin{tikzpicture}[->, >=stealth, node distance=1cm]
        \node (X1) at (0,0) {$X_1$};
        \node (X2) at (1,0) {$X_2$};
        \node (X3) at (2,0) {$X_3$};

        \draw (X2) -- (X1);
        \draw (X3) -- (X2);
    \end{tikzpicture}
    \caption{Two Markov equivalent DAGs}
    \label{MarkovEquivalence}
\end{figure}

\begin{figure*}[t]
    \centering
    \includegraphics[width=0.97\textwidth]{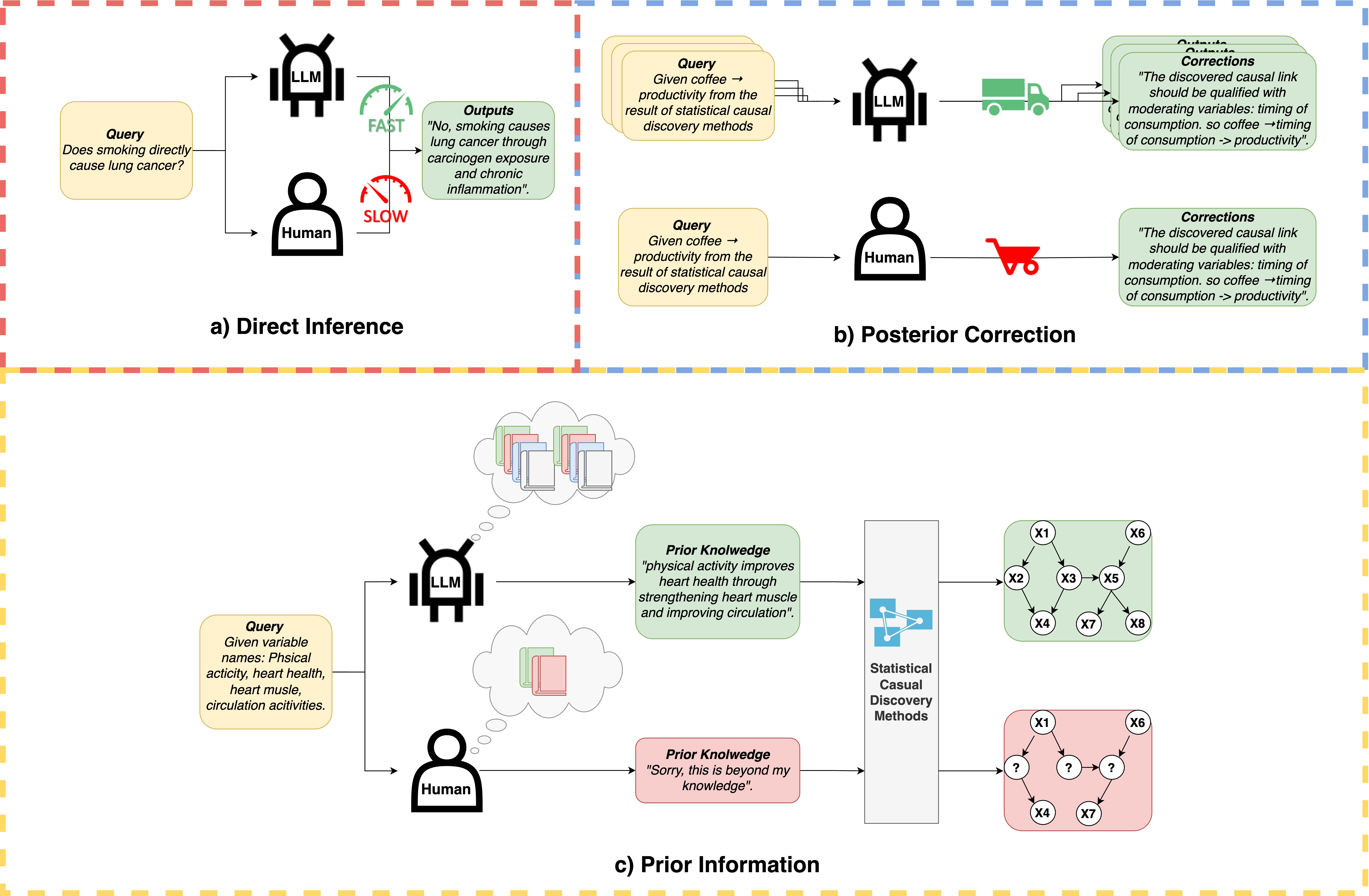}  % 90% of text width
    \caption{Three distinct approaches for applying LLMs in causal discovery: (a) Direct causal inference without observational data, (b) Post refinement of statistically derived causal structures, and (c) Integration of prior knowledge into traditional statistical methods. The figure highlights the increasing automation and precision of causal discovery through LLMs, reducing the need for manual expert input.}
    \label{fig:causal_discovery}
\end{figure*}

\noindent \textbf{Structure Learning}: Structure learning aims at estimating a graph that captures dependence relationships in data~\cite{drton2017structure}.  While any probability measure $\mathbb{P}(\mathbf{X})$ can be factored as:
\begin{equation}
\mathbb{P}(\mathbf{X}) = \mathbb{P}(X_1)\mathbb{P}(X_2|X_1) \cdots \mathbb{P}(X_n|X_1, \dots, X_{n-1}),
\end{equation}
this complete graph representation introduces unnecessary dependencies. The goal is to discover a minimal structure where directed edges $X_i \to X_j$ represent only necessary conditional dependencies~\cite{pearl2009causal}.

A fundamental limitation is that different Directed Acyclic Graphs (DAGs) can encode identical conditional independence relations, forming Markov equivalence classes. For example, given variables $X_1$, $X_2$, and $X_3$ with the following conditional independence relations:
\begin{equation}\label{eq:independence_relation}
X_1 \not\!\perp X_2, \quad X_2 \not\!\perp X_3, \quad X_1 \not\!\perp X_3, \quad X_1 \perp X_3 \mid X_2,
\end{equation}
multiple network structures (Figure~\ref{MarkovEquivalence}) can represent these relationships, demonstrating that causal direction cannot be uniquely determined from observational data alone.

\label{knowledge driven}

\begin{table*}[t]
\centering
\caption{A practical example of prompts with respect to various LLM causal discovery frameworks.}
\label{tab:example}
\begin{tabular}{p{3.5cm}|p{14cm}}
\hline
\textbf{Tasks} & \textbf{Prompt} \\ 
\hline
Pairwise Discovery & 
"Which is more likely to be true: (A) lung cancer causes cigarette smoking, or (B) cigarette smoking causes lung cancer?" \\ 
\hline
Conditional Independence Set Test & 
As an expert in a specific field, you're asked to assess the statistical independence between two variables, potentially conditioned on another variable set. Your response, based on theoretical knowledge, should be a binary guess (YES or NO) and the probability of its correctness, formatted as: [ANSWER (PROBABILITY\%)]. For example, [YES (70\%)] or [NO (30\%)]. \\
\hline
Causal Validation &
Given a statistical correlation between variables A and B, and their relationship with other variables in the system, determine if we can validly conclude that A causes B. Please provide your reasoning step by step and conclude with either 'Valid' or 'Invalid' for this causal inference. \\
\hline
Full Graph Discovery & 
As a domain expert, analyze cause-and-effect relationships among variables with given abbreviations and values. Interpret each variable and present the causal relationships as a directed graph, using edges to denote direct causality, e.g., \(x_{i1} \rightarrow x_{j1}\), \(\ldots\), \(x_{im} \rightarrow x_{jm}\). \\
\hline
\end{tabular}
\end{table*}
\noindent \subsection{Causal Discovery}

A causal graph is a Directed Acyclic Graph (DAG) where edges represent direct causal relationships—an edge $X_i \to X_j$ indicates that $X_i$ is a direct cause of $X_j$. Building upon this foundation, \textbf{Causal Discovery (CD)} is the systematic process of uncovering these causal relationships from observational data~\cite{glymour2019review}. When all common causes are included in the observed variables, the problem typically reduces to learning Bayesian Network structures with causal interpretations, serving as the foundation for various downstream applications like effect inference \cite{kuang2020causal} and prediction \cite{chu2023task}.

\noindent \textbf{Structural Causal Model (SCM).} A Structural Causal Model (SCM) provides a formal framework for representing causal systems \cite{pearl2009causal}. An SCM is defined as $M = (V, U, F, P)$, where $V$ represents the set of observable variables, $U$ is the set of unobservable (or latent) variables, $F$ is the collection of functions describing the causal mechanisms, and $P$ is the probability distribution over the variables. Each variable $X_i \in V$ is determined by a function $f_i \in F$ of its direct causes (parents) $Pa(X_i)$ and an independent noise term $U_i \in U$:
\begin{equation}
    X_i = f_i(Pa(X_i), U_i)
\end{equation}
 If SCM is employed, the goal of causal discovery is to recover these structural components from data, particularly identifying the parent sets $Pa(X_i)$ for each variable $X_i$, which defines the underlying causal graph structure. While discovering the exact functional relationships $f_i$ is challenging, methods typically focus on learning this causal graph structure, which represents the dependencies between variables.

Statistical approaches to causal discovery have evolved into two main methodologies. \textbf{Constraint-based methods}, exemplified by the PC Algorithm~\cite{spirtes2001causation}, start with a complete undirected graph and systematically test conditional independence relationships ($X \perp Y|Z$) to remove edges and orient directions, operating under assumptions of causal sufficiency and faithfulness \cite{glymour2019review}. \textbf{Score-based methods} search through possible DAGs by optimizing scoring functions (such as BIC or BGe) that balance model fit against complexity, with recent advances like NOTEARS~\cite{zheng2018dagstearscontinuousoptimization} reformulating graph learning as a continuous, differentiable optimization problem.

\subsection{Tasks and Evaluation Metrics} Causal discovery with LLMs addresses two distinct tasks with different complexity levels. \textbf{Causal Order Predictions}, or pairwise discovery, involves evaluating direct causal relationships between variable pairs through natural language queries. As each pair represents an independent binary classification problem, performance is evaluated using standard metrics like accuracy and F1-score. In contrast, \textbf{Full Graph Discovery} aims to construct complete causal networks, where edge decisions are interdependent, making it an NP-complete problem. This increased complexity necessitates different evaluation approaches, primarily the Structural Hamming Distance (SHD)  and Normalized Hamming Distance (NHD) \cite{Tsamardinos2006MaxMinHC}, which measure the number of edge operations needed to transform the learned graph into the true causal structure.

\subsection{Large Language Model}

Large Language Models (LLMs) such as GPT-4 and LLaMA have revolutionized natural language processing through their advanced transformer architecture , demonstrating remarkable capabilities in reasoning, knowledge acquisition, and cross-domain generalization \cite{zhao2023survey}. Beyond basic language tasks, these models excel in complex reasoning and knowledge integration, making them particularly suitable for causal discovery tasks. Their ability to perform step-by-step deductive reasoning and leverage extensive pre-trained knowledge across various domains enables them to understand and identify causal relationships in different contexts \cite{wei2022chain}.

\section{LLMs for Causal Discovery}
\label{llm4causal}

This section explores how LLMs can succeed the role of human domain experts, as shown in Figure \ref{fig:causal_discovery}, discussing three primary approaches to incorporating LLMs in causal discovery : (a) direct inference, (b) posterior refinement on derived causal structures, and (c) knowledge integration as prior for generating causa structure, and Table~\ref{tab:example} provides practical examples of prompts used in these various approaches.

\subsection{LLMs as Direct Inference}
\label{DI}
Direct causal discovery leverages LLMs' extensive knowledge acquired during pre-training to serve as automated domain experts capable of reasoning about causal relationships. Unlike traditional methods that rely solely on statistical patterns or require extensive human expert consultation, LLMs can utilize their broad understanding of domain concepts, scientific principles, and real-world relationships to infer causality at scale. The fundamental setting involves providing LLMs with meta-data such as the descriptive texts $T = \{t_1, t_2, \ldots, t_n\}$ for variables $X = \{x_1, x_2, \ldots, x_n\}$. By comprehending these descriptions and applying their learned knowledge, LLMs identify causal statements denoted as $S = \{(x_i, x_j)\}$, where $(x_i, x_j)$ indicates that $x_i$ causes $x_j$. This approach effectively transforms LLMs into scalable meta-data experts who can reason about causality with both breadth and precision.

Two primary approaches have emerged in this direction. First, \textbf{causal order prediction}, pioneered by \cite{willig2022foundation} and advanced by \cite{kıcıman2023causal}, focuses on determining pairwise causal relationships through direct LLM queries. Using chain-of-thought prompting \cite{wei2022chain}, these methods systematically extract causal mechanisms from LLMs' pre-trained knowledge. Second, \textbf{complete and partial causal graph discovery} methods aim to identify broader causal structures through iterative pairwise discovery \cite{long2024large,kıcıman2023causal}, though this naturally introduces computational challenges scaling quadratically with the number of variables. To address these efficiency barriers, \cite{jiralerspong2024efficient} reduced computational complexity from $\mathcal{O}(n^2)$ to $\mathcal{O}(n)$ through a structured three-phase process: root cause identification, relationship expansion, and logical consistency verification. For high-dimensional settings, \cite{sokolov24} developed a scalable solution using hierarchical clustering based on semantic similarity, efficiently discovering causal relationships by first analyzing within-cluster connections before determining inter-cluster causality. To enhance the reliability of these approaches, researchers have pursued several complementary directions: systematic verification to mitigate hallucination \cite{ji2023survey}, specialized fine-tuning for causal reasoning \cite{le2024multiagentcausaldiscoveryusing}, and structured prompting frameworks for consistent causal extraction \cite{zhang2024causalgraphdiscoveryretrievalaugmented,vashishtha2023causalinferenceusingllmguided}.

\noindent \textbf{Remark:} Despite these advances, LLMs alone show limitations in causal discovery performance, motivating new methodological advancement on approaches that combine LLM capabilities with traditional statistical causal discovery methods.

\begin{figure}[t]
\centering
\includegraphics[width=\columnwidth]{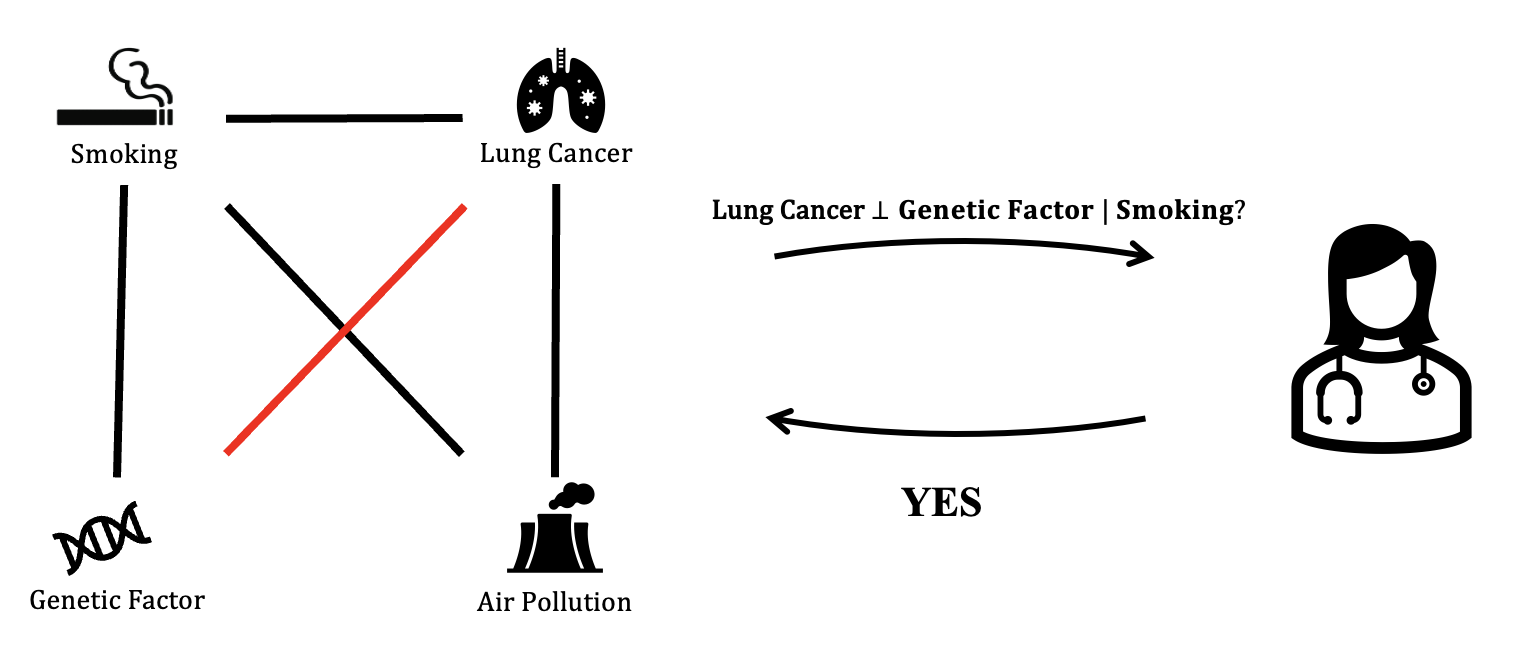}
\caption{General Workflow of Applying LLM on Conditional Independence Tests to Solve Constraint-Based Discovery Tasks.}
\label{fig:CI}
\end{figure}

\subsection{LLMs as Posterior Correction} 
\label{Posterior}
The first way LLMs can be used in conjunction with traditional methods is that LLMs can serve as a expert judge by correcting and refining the learned causal structures from the traditional statistical causal discovery (SCD) methods based on contextual reasoning or additional data.

Recall that most of constraint-based and score-based methods can only identify a BN up to its Markov equivalence class. Given the set of all conditional independence relations, \cite{long2023causal} introduces a method that uses LLM as an imperfect expert to progressively reduce the number of possible causal structures within the Markov equivalence class, while controlling the risk of misorienting edges. However, its application to larger datasets remains unproven and is likely hindered by the approach's complexity and computational demands.

Instead of deriving the true causal graph beyond its Markov equivalence class, the Iterative LLM Supervised CSL Framework (ILS-CSL) by \cite{ban2023causal} refines a partially learned DAG by leveraging LLM feedback iteratively to correct edge orientations. This approach efficiently integrates expert knowledge while avoiding exhaustive pairwise queries, ensuring a more accurate and robust causal structure without requiring full causal discovery from scratch. 

\cite{takayama2024integrating} introduced a framework in which large language models (LLMs) function both as post hoc refiners and prior knowledge generators. Initially, the raw adjacency matrix \(\hat{G}_0\), derived from certain SCD methods without prior knowledge, is input into the LLMs. For each potential edge, the LLM is queried multiple times using pairwise prompts, with each response being a binary decision (i.e., "yes" or "no"). The probability matrix \(P\) is then constructed by aggregating these responses, and is subsequently converted into a deterministic prior knowledge matrix \(\hat{G}\) based on predefined thresholds. Finally, \(\hat{G}\) is incorporated into the original SCD method to infer the final causal graph. While this framework is primarily designed for LiNGAM, it is applicable to any causal discovery algorithm that relies on a matrix of pairwise edge scores.

\cite{abdulaal2024causal} introduced the Causal Modeling Agent (CMA), a novel framework for causal discovery that synergistically combines the metadata-based reasoning of LLMs with the data-driven power of Deep Structural Causal Models (DSCMs).  CMA employs an LLM to propose an initial causal graph, which then informs the fitting of a DSCM to the data.  The framework iteratively refines this graph in global and local phases, again using the LLM as both a provider of prior knowledge and a critic of the model's output, enabling the discovery of causal relationships in complex, multi-modal data.  Unlike purely constraint-based or scoring-based methods, CMA integrates aspects of both while leveraging LLMs.  Furthermore, it can generate chain graphs to account for unmeasured confounding and has demonstrated state-of-the-art performance on datasets like Arctic Sea \cite{huang2021benchmarking} and on synthetic data designed to prevent data leakage.

\subsection{LLMs as Prior Knowledge} 
\label{Prior}

Similarly, LLMs can also be used in conjunction with traditional methods to provide source of prior knowledge by leveraging meta-data extracted from textual descriptions and domain-specific information. In \cite{ban2023query}, a set of variables \textbf{X} along with their descriptive texts \textbf{T} are provided as input to an LLM, which performs causal discovery by identifying direct relationships after comprehending the semantic meaning of each variable. Traditionally, the incorporation of prior knowledge into CD procedures follows two primary approaches: the hard constraint method and the soft constraint method. The hard constraint approach strictly enforces prior knowledge by eliminating edges in BNs that conflict with constraints derived from traditional causal discovery algorithms. However, this method lacks flexibility, as any spurious prior assumptions cannot be corrected during the learning process, potentially leading to erroneous causal structures. \cite{chen2023mitigatingpriorerrorscausal} provides a systematic approach for detecting and correcting potentially erroneous prior knowledge derived from LLMs, thereby enhancing the reliability of utilizing LLMs for hard prior constraints.

In contrast, the soft constraint approach, though implemented through varying methodologies, aims to integrate prior knowledge in a fault-tolerant manner. \cite{ban2023query} integrates LLM-derived prior into the usual scoring functions, such as BDeu or BIC, as a regularization term, thereby allowing flexibility in cases where LLM priors may be inconsistent with observed data. Alternatively, \cite{darvariu2024large} introduces a probabilistic prior framework, where the LLM-derived priors consist of probabilities of the existence and direction of each edge in the causal graph, which are then incorporated into some traditional CD algorithms. Notably, the proposed approach is compatible with any causal discovery algorithm that relies on a matrix of pairwise edge scores, including LiNGAM and NOTEARS. Given the demonstrated effectiveness of integrating LLMs into differentiable causal discovery algorithms within this probabilistic framework, recent work by \cite{kampani2024llminitializeddifferentiablecausaldiscovery} further extends this paradigm by utilizing LLMs to initialize the continuous optimization process.

A novel application of LLMs as providers of prior information lies in conditional independence testing, a cornerstone of constraint-based causal discovery.  Rather than using traditional statistical tests, LLMs can be queried with natural language prompts representing conditional independence relationships, effectively serving as an oracle.  As shown in Figure~\ref{fig:CI}, this enables constainted based algorithms like PC algorithm \cite{spirtes2001causation} to leverage LLMs for guidance in causal graph construction. The chatPC method~\cite{cohrs2023large} exemplifies this approach, integrating LLMs with the PC algorithm by transforming conditional independence tests (e.g., "Is X independent of Y given Z?") into natural language prompts.  The LLM's responses then guide the PC algorithm's edge removal process.  This work evaluates LLM performance on such queries, proposes a statistical aggregation method to combine multiple LLM responses for increased robustness, and analyzes the resulting causal graphs.  Research indicates that LLMs tend to be more conservative in their independence judgments than human experts, yet still demonstrate evidence of causal reasoning.

\noindent \textbf{Remark: } This integration of LLMs with SCD, as discussed in the previous two subsections, offers the potential for overcoming limitations of traditional causal discovery and purely knowledge-based LLM-driven causal discovery, paving the way for more automated, scalable, and interpretable causal graph construction.

\section{Evaluations and Applications}
\label{Eval}
\begin{table*}[htbp]
\centering
\caption{Summary of Benchmark Datasets for Evaluating Causal Discovery Tasks}
\label{tab:data}
\renewcommand{\arraystretch}{0.9}  % Slightly reduced row height
{\footnotesize  % Added footnotesize for entire table
\begin{tabularx}{\textwidth}{lcccccccc}
\toprule
\textbf{Dataset Name} & \textbf{Work} & \textbf{Pair} & \textbf{Full} & \textbf{Novel} & \textbf{Domain} & \textbf{Avg. Nodes} & \textbf{Avg. Edges} & \textbf{Num. Graphs} \\
\midrule
% Single Graph Datasets (Chronological Order)
Asia & \cite{pearl1988local} & \checkmark & \checkmark & \checkmark & Medical & 8 & 8 & -- \\
Insurance & \cite{binder1997adaptive} & \checkmark & \checkmark & \checkmark & Business & 27 & 52 & -- \\
CauseNet & \cite{heindorf2020causenet} & \checkmark & $\times$ & $\times$ & Web/Mixed & 12.2M & 11.6M & -- \\
Arctic Sea & \cite{huang2021benchmarking} & \checkmark & \checkmark & $\times$ & Climate & 12 & 42 & -- \\
Neuropathic & \cite{tu2023causaldiscovery} & $\times$ & \checkmark & $\times$ & Medical & 222 & 770 & -- \\
Sangiovese & \cite{kıcıman2023causal} & $\times$ & \checkmark & $\times$ & Agriculture & 15 & 55 & -- \\
Tübingen & \cite{kıcıman2023causal} & \checkmark & $\times$ & \checkmark & Mixed Science & 222 & 770 & -- \\
Alzheimer & \cite{abdulaal2024causal} & \checkmark & \checkmark & \checkmark & Medical & 11 & 19 & -- \\
\midrule
\multicolumn{9}{c}{\textbf{Multi-Graph Benchmark Datasets}} \\
\midrule
CLADDER & \cite{jin2023cladder} & \checkmark & \checkmark & \checkmark & Mixed & 3.52 & 3.38 & 10,112 \\
CORR2CAUSE & \cite{jin2023large} & \checkmark & \checkmark & \checkmark & Mixed & 2-6 & 8.60 & 207,972 \\
AppleGastronome & \cite{liu2024discoveryhiddenworldlarge}  & \checkmark & \checkmark & \checkmark & Food & 6 & 5 & 200 \\
CausalBench & \cite{zhou2024causalbenchcomprehensivebenchmarkcausal} & \checkmark & \checkmark & \checkmark & Mixed & 2-109 & 11.6M & 15 \\
\bottomrule
\end{tabularx}
}  % End of footnotesize
\vspace{1mm}  % Added small space before the note
{\scriptsize Note: Pair = Pairwise Discovery; Full = Full Graph Discovery; Novel = Involves a stimulator to regenerate data to avoid data leakage}  % Made note even smaller
\end{table*}

\subsection{Benchmarks and Datasets}
\label{Data}

Table \ref{tab:data} provides a comprehensive overview of benchmark datasets used to evaluate LLMs' causal reasoning capabilities. For each dataset, the table indicates  (1) if for determining causal relationships between variable pairs, (2) full graph reconstruction, and (3) novel reasoning scenarios (testing on previously unseen causal patterns). The table also details key characteristics including the average number of nodes and edges per graph, along with the total number of graphs in the collection. Multi-graph benchmark datasets, such as CausalBench \cite{zhou2024causalbenchcomprehensivebenchmarkcausal}, are particularly noteworthy as they incorporate established causal networks dataset commonly tested in the literature, such as Asia \cite{pearl1988local} and Insurance \cite{binder1997adaptive}, offering evaluation across diverse graph sizes and domains. Among these benchmarks, CORR2CAUSE \cite{jin2023cladder} addresses a crucial aspect of causal reasoning: the ability to differentiate causation from correlation and can be further used for fine-tuning LLMs to enhance their causal inference capabilities from identifying purely correlational statements.

\subsection{Applications}
\label{Apps}
Causal discovery has been used as a crucial tool across numerous real-world domains, with LLM-based methods significantly expanding its capabilities and applications. For instance, \cite{gkountouras2024languageagentsmeetcausality} introduced a "causal world model" framework, connecting causal variables to natural language to improve reasoning in complex environments.  To address the challenge of ill-defined high-level variables often found in real-world observational data, \cite{liu2024discoveryhiddenworldlarge} enables LLMs to propose such variables, effectively extending causal discovery to unstructured data.  These and other advances have facilitated LLM-guided causal discovery in fields like medicine \cite{tu2023causaldiscovery,cohrs2025large}, finance \cite{sokolov24}, genetics \cite{afonja2024llm4grndiscoveringcausalgene}, and health informatics \cite{10.1145/3698587.3701384}.  Further research explores LLM-driven causal discovery in multi-agent systems \cite{abdulaal2024causal,jiang2024llmcausal} and multi-modal data integration \cite{shen2024exploringmultimodalintegrationtoolaugmented}, leveraging the richness of multi-modal data to provide additional information to better capture the complexity of real-world systems.

\section{Challenges and Visions}
\label{Future}
\paragraph{Unified Evaluation and Fair Comparison}

A significant challenge in LLM-enhanced causal discovery is the lack of standardized evaluation protocols and comprehensive comparative analyses. While many methods have been proposed, current research often lacks systematic comparison between different LLM-based approaches, making it difficult to establish true state-of-the-art performance. To address these challenges, we propose that researchers should evaluate their methods using both synthetic datasets (to avoid potential data leakage from LLM pre-training) and established benchmarks (for fair comparison). When computational constraints necessitate using subsets of larger graphs, especially in full graph discovery tasks, the community should establish standardized sampling procedures. Furthermore, we advocate for a comprehensive evaluation framework that incorporates multiple performance metrics beyond accuracy, including sub-graph and domain-specific metrics and computational efficiency, while maintaining clear documentation of prompt engineering strategies and LLM configurations to enhance reproducibility.

\paragraph{Domain-Specific LLMs and Expert Agents for Causal Discovery}

While current approaches primarily utilize general-purpose LLMs for causal discovery, due to the applications of causal discovery such as healthcare and economics, we envision significant potential in developing domain-specialized models and expert agents that can better capture field-specific causal relationships. Future work should focus on creating domain-expert agents through strategic integration of LLMs with field-specific knowledge bases and specialized causal reasoning modules. These expert systems could be enhanced through domain-specific Retrieval Augmented Generation (RAG), incorporating scientific literature, domain ontologies, and expert-curated databases. For instance, in molecular biology, such agents could access pathway databases and protein interaction networks, while in climate science, they could integrate physics-based models and environmental data. Furthermore, these expert systems could benefit from continuous learning mechanisms that update their knowledge based on new research findings and domain-specific discoveries, ensuring their causal reasoning remains current and scientifically rigorous. 
\paragraph{LLM for SCM Diagnosis}

Besides graph discovery, Future research should expand LLMs' role beyond simply identifying causal relationships to understanding and verifying the underlying properties of Structural Causal Models (SCMs). While current approaches primarily use LLMs to detect the presence of causal links \cite{jin2023large}, there's significant potential for LLMs to verify crucial aspects like the nature of causal relationships (linear vs. nonlinear), functional forms of causal effects, and characteristics of noise distributions. This verification is particularly important since the effectiveness of traditional causal discovery methods often depends on specific assumptions - for instance, DirectLiNGAM requires linear relationships \cite{shimizu2011directlingam}, and BIC scoring becomes less reliable with nonlinear causal effects\cite{peters2017elements}. LLMs could potentially leverage their natural language understanding capabilities to interpret domain knowledge about the expected nature of causal relationships, helping to select appropriate causal discovery algorithms and validate their assumptions. This deeper integration of LLMs into the causal discovery pipeline could potentially improve the reliability and applicability of causal discovery methods across different scenarios and domains.

\paragraph{Explanability and Interpretbility}

Despite advancements in statistical causal discovery (SCD), the capability of Large Language Models (LLMs) to perform genuine causal discovery tasks remains questionable. Many research studies how explored what potential factors contribute to the causal reasoning of LLMs \cite{ze2023causal,feng2024pretrainingcorporalargelanguage}, suggesting that LLMs are like parrots in that they simply recite the causal knowledge embedded in the data. The analysis discussed in \cite{jin2023cladder} provides evidence that current LLMs perform even more poorly than expected, functioning as weak 'causal parrots' that merely recite embedded causal knowledge without deeper understanding. Future research in LLM-based causal discovery should focus on three key areas: (1) developing interpretability methods to analyze how LLMs process causal relationships, including attention pattern analysis and systematic probing studies , (2) investigating the relationship between pre-training data distributions and causal learning capabilities, and (3) creating evaluation frameworks that can distinguish between genuine causal reasoning and spurious correlations. This understanding could advance LLM development bu designing architectures that prioritize genuine causal reasoning over pattern recognition, examining internal representations during intervention-based tasks, and investigating how models differentiate causation from correlation.

\section{Conclusion}
The integration of Large Language Models (LLMs) with causal discovery marks a transformative advancement in artificial intelligence. This survey has explored how LLMs are revolutionizing traditional causal discovery through direct inference, knowledge integration, and structural refinement. While LLMs offer significant potential for democratizing causal discovery and reducing dependence on domain experts, challenges remain, including the need for robust evaluation frameworks and ensuring the trustworthiness of LLM-driven causal reasoning.  Addressing these challenges through targeted research is crucial for future progress.  As these methods mature, we expect the synergy between LLMs and causal discovery to yield increasingly sophisticated and reliable tools for understanding causal relationships across diverse domains, ultimately accelerating scientific discovery and improving decision-making.
\newpage

\appendix

\bibliographystyle{named}
\bibliography{ijcai25}

\end{document}